\documentclass[10pt,twocolumn,twoside]{IEEEtran}
\usepackage[utf8]{inputenc}         
\usepackage[english]{babel}
\usepackage{multirow}

\usepackage{multicol}

\usepackage{soul}

\usepackage[normalem]{ulem}

\usepackage{cite}

\ifCLASSINFOpdf
  \usepackage[pdftex]{graphicx}
  \usepackage{epstopdf}
  \graphicspath{{./image/}}
\else
\fi
\usepackage[cmex10]{amsmath}
\usepackage{amssymb}
\usepackage{array}
\usepackage{rotating}

\usepackage{makecell}
\usepackage{subcaption}
\usepackage{textcomp}
\usepackage{url}

\usepackage{lipsum}

\usepackage{url}

\usepackage{multirow}
\usepackage{balance}
\usepackage{cite}
\usepackage{capt-of}
\usepackage[cmex10]{amsmath}
\usepackage{amssymb}

\usepackage{tabularx}
\usepackage{textcomp}
\usepackage{bm}
\usepackage{calc}

\DeclareMathOperator*{\argmax}{arg\,max}
\usepackage{xcolor,colortbl}

\usepackage{tabu}
\usepackage{mdframed}

\makeatletter
\def\blfootnote{\xdef\@thefnmark{}\@footnotetext}
\makeatother

\usepackage[ruled,lined]{algorithm2e}

\begin{document}
\bstctlcite{IEEEexample:BSTcontrol}
%
\title{Joint Classification and Prediction CNN Framework for Automatic Sleep Stage Classification}

%
%
%

\author{Huy~Phan$^*$,~\IEEEmembership{Member,~IEEE,}
				Fernando~Andreotti,~\IEEEmembership{Member,~IEEE,}
				Navin~Cooray,~\IEEEmembership{Student Member,~IEEE,}
				Oliver~Y.~Ch\'{e}n,~\IEEEmembership{Student Member,~IEEE,}
        and~Maarten~De~Vos
\thanks{H. Phan, F. Andreotti, N. Cooray, O. Y. Ch\'{e}n, and M. De Vos are with the Institute of Biomedical Engineering, University of Oxford, Oxford OX3 7DQ, United Kingdom.}
\thanks{$^*$Corresponding author: {\tt\footnotesize huy.phan@eng.ox.ac.uk}}}

%
%

\markboth{This Article Has Been Published in IEEE Transactions on Biomedical Engineering}
{This Article Has Been Published in IEEE Transactions on Biomedical Engineering}
%



\maketitle

\begin{abstract}

Correctly identifying sleep stages is important in diagnosing and treating sleep disorders. This work proposes a joint classification-and-prediction framework based on convolutional neural networks (CNNs) for automatic sleep staging, and, subsequently, introduces a simple yet efficient CNN architecture to power the framework\footnote{\footnotesize The source code and the relevant experimental setup are available at \\ \url{http://github.com/pquochuy/MultitaskSleepNet} for reproducibility.}. 
Given a single input epoch, the novel framework jointly determines its label (classification) and its neighboring epochs' labels (prediction) in the contextual output. While the proposed framework is orthogonal to the widely adopted classification schemes, which take one or multiple epochs as contextual inputs and produce a single classification decision on the target epoch, we demonstrate its advantages in several ways. First, it leverages the dependency among consecutive sleep epochs while surpassing the problems experienced with the common classification schemes. Second, even with a single model, the framework has the capacity to produce multiple decisions, which are essential in obtaining a good performance as in ensemble-of-models methods, with very little induced computational overhead. Probabilistic aggregation techniques are then proposed to leverage the availability of multiple decisions. 
To illustrate the efficacy of the proposed framework, we conducted experiments on two public datasets: Sleep-EDF Expanded (Sleep-EDF), which consists of 20 subjects, and Montreal Archive of Sleep Studies (MASS) dataset, which consists of 200 subjects. The proposed framework yields an overall classification accuracy of 82.3\% and 83.6\%, respectively. 
We also show that the proposed framework not only is superior to the baselines based on the common classification schemes but also outperforms existing deep-learning approaches. 
To our knowledge, this is the first work going beyond the standard single-output classification to consider multitask neural networks for automatic sleep staging. This framework provides avenues for further studies of different neural-network architectures for automatic sleep staging.
\end{abstract}


%
\IEEEpeerreviewmaketitle

\section{Introduction}
\label{sec:intro}

\blfootnote{\footnotesize DOI: 10.1109/TBME.2018.2872652}
Identifying the sleep stages from overnight Polysomnography (PSG) recordings plays an important role in diagnosing and treating sleep disorders, which affects millions of people \cite{Krieger2017,Redmond2006}.
Traditionally, this task has been done manually by experts via visual inspection which is tedious, time-consuming, and is prone to subjective error. Automatic sleep stage classification \cite{Aboalayon2016}, that performs as well as manual scoring, can help to ease this task tremendously, therefore facilitating home monitoring of sleep disorders \cite{Kelly2012}.

\begin{figure} [!t]
	\centering
	\includegraphics[width=0.9\linewidth]{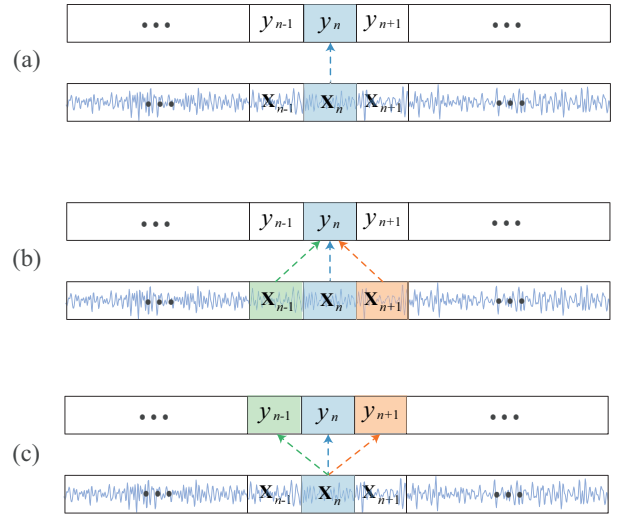}
	\vspace{0.2cm}
	\caption{Illustration of (a) the standard classification approach, (b) the common classification approach with the contextual input of three epochs, and (c) the joint classification and prediction with the contextual output of three epochs proposed in this work.}
	\label{fig:contextualinput_vs_output}
\end{figure}

The guiding principle of automatic sleep staging is to split the signal into a sequence of epochs, each of which is usually 30 seconds long, and the classification is then performed epoch-by-epoch. 
In order to uncover a sleep stage at each epoch, proper features need to be derived from the signal, such as electroencephalography (EEG). Traditionally, many features have been designed based on prior knowledge of sleep. These hand-crafted features range from time-domain features \cite{Krakovska2011, Koley2012, Redmond2006} to frequency-domain features \cite{Phan2013, Susmakova2008, Koley2012, Fell1996}, via features derived from nonlinear processes \cite{Kim2000, Lee2012, Susmakova2008, Zhang2001}. Using these features, the classification goal is often accomplished by conventional machine learning algorithms, such as Support Vector Machine (SVM) \cite{Alickovic2018,Koley2012}, $k$-nearest neighbors ($k$-NN) \cite{Phan2013}, Random Forests \cite{Memar2018, Boostania2017,Imtiaz2015b}.

The advent of deep learning and its astonishing progress in numerous domains have stimulated interest in applying them for automatic sleep staging. The power of deep networks lies in their great capability of automatic feature learning from data, thus avoiding the reliance on hand-crafted features. Significant progress on results obtained from different sleep staging benchmark using various deep learning techniques have been reported \cite{Stephansen2017, Mikkelsen2018, Zhang2017, Tsinalis2016, Tsinalis2016b, Supratak2017, Dong2017, Phan2018c, Phan2018d}, mirroring a relentless trend where learned features ultimately outperform and displace long-used hand-crafted features. CNN \cite{LeCun2015,Lecun1989}, the cornerstone of deep learning techniques, has been frequently employed for the task \cite{Mikkelsen2018, Zhang2017, Tsinalis2016}. The weight sharing mechanism at the convolutional layers forces the shift-invariance of the learned features and greatly reduces the model's complexity, consequently leading to improvement of the model's generalization \cite{LeCun2015}. Other network variants, such as Deep Belief Networks (DBNs) \cite{Laengkvist2012}, Auto-encoder \cite{Tsinalis2016b}, Deep Neural Networks (DNNs) \cite{Dong2017}, have also been explored. Moreover, Recurrent Neural Networks (RNNs), e.g. Long Short-Term Memory (LTSM) \cite{Hochreiter1997}, which are capable of sequential modelling, have been found efficient in capturing long-term sleep stage transition and are usually utilized to complement other network types, such as CNNs \cite{Supratak2017,Stephansen2017} and DNNs \cite{Dong2017}. Standalone RNNs have also been exploited for learning sequential features of sleep \cite{Phan2018d,Koch2018a,Koch2018b}. The classification is usually performed therein by the networks in an end-to-end fashion \cite{Mikkelsen2018, Zhang2017, Tsinalis2016}; a separate classifier, such as SVM, can be used alternatively \cite{Phan2018d,Ansari2018}.

\section{Motivation and Contributions}
\subsection{Motivation}
Sleep is a temporal process with slow stage transitions, implying continuity of sleep stages and strong dependency between consecutive epochs \cite{Iber2007, Liang2011,Sousa2015}. For instance, out of 228,870 epochs in the entire MASS dataset \cite{Oreilly2014} used in this work,  $83.3\%$ pairs of adjacent epochs have the same label. The ratio is still as high as $79.3\%$, when two epochs are separated by one epoch. This nature of sleep has inspired a widely adopted practice in neural-network-based sleep staging systems, namely the use of \emph{contextual input} that augments a target epoch by its surrounding epochs (\emph{many-to-one}) in the classification task \cite{Stephansen2017, Mikkelsen2018, Tsinalis2016, Supratak2017}. Common input context size is of three and five epochs \cite{Chambon2018,Mikkelsen2018,Tsinalis2016, Tsinalis2016b}. This classification scheme can also be interpreted as an extension of the standard classification setup,  i.e. determining the sleep stage corresponding to a single epoch of input signals (\emph{one-to-one}) \cite{Phan2018c, Phan2018d, Andreotti2018}. Figure~\ref{fig:contextualinput_vs_output}~(b) provides a schematic presentation of contextual input of three epochs in comparison with the standard one-to-one classification approach in Figure~\ref{fig:contextualinput_vs_output}~(a). While multiple-epoch input  does not always provide performance gains, as shown in our experiments, it poses a problem of inherent \emph{modelling ambiguity}. That is, when training a network with contextual input, such as three epochs illustrated in Figure \ref{fig:contextualinput_vs_output} (b), it remains unclear whether the network is truly modelling the class distribution of the target epoch at the center or that of the left and right neighbor. In our experiment, such a network (i.e. the many-to-one baseline in Section \ref{ssec:baseline}) achieves an accuracy of $82.1\%$ in determining the labels of the center epochs. However, when aligning the network output with those labels of the left and right neighbor, the accuracy is just marginally lower, reaching $81.1\%$ and $80.8\%$, respectively. Last but not least, the contextual input causes the network's computational complexity increase at a linear scale due to the enlarged input size.

\begin{figure*} [!t]
	\centering
	\includegraphics[width=0.85\linewidth]{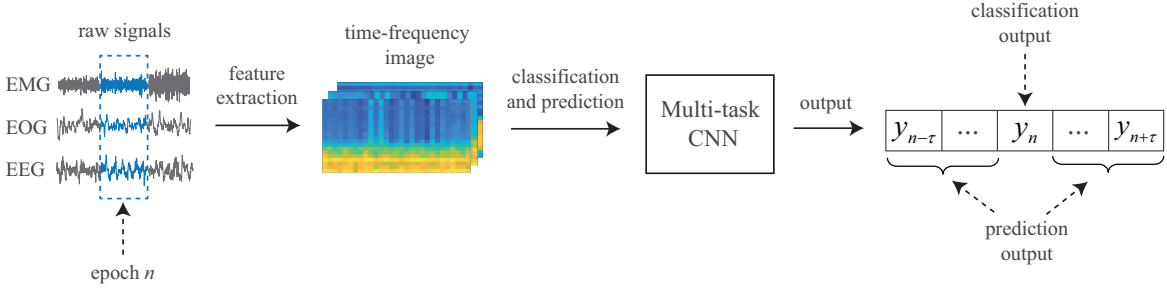}
	\vspace{-0.1cm}
	\caption{Overview of the proposed joint classification and prediction framework.}
	\label{fig:overview}
	\vspace{-0.2cm}
\end{figure*}

In this work, we formulate sleep staging as a joint classification and prediction problem. In other words, this is equivalent to a \emph{one-to-many} problem, which is an extension of the standard one-to-one classification scheme while being orthogonal to the common many-to-one classification scheme. With this new formulation, given a single target epoch as input, our objective is to simultaneously determine its label (classification) and its neighboring epochs' labels (prediction) in the \emph{contextual output}, as demonstrated in Figure~\ref{fig:contextualinput_vs_output}~(c). By classification, we mean determining the label of an epoch given its information. In contrast, prediction implies determining the label of an epoch without knowing its information. The rationale behind this idea is that, given the strong dependency of consecutive epochs, using information of an epoch, we should be able to infer the label of its neighbors. The major benefit of the joint classification and prediction formulation are two-fold. First, with the single-epoch input, the employed model does not experience the modelling ambiguity and the computational overhead induced by the large contextual input as previously discussed. Second, the employed model can
produce an ensemble of decisions, which is the key in our obtained state-of-the-art performance, with a negligible induced computational cost. Ensemble of models \cite{Hinton2015,Dietterich2000}, a well-established method to improve the performance of a machine learning algorithm, has been found generalizable to automatic sleep staging, evidenced by conventional methods \cite{Alickovic2018, Bhuiyan2016, Koley2012} and recently developed deep neural networks \cite{Stephansen2017}. However, building many different models on the same data for model fusion is cumbersome and costly. Opposing to ensemble of models \cite{Stephansen2017, Alickovic2018, Bhuiyan2016, Koley2012}, in our joint classification and prediction formulation, the ensemble of decisions is produced with a \emph{single} multi-task model. Afterwards, an aggregation method can be used to fuse the ensemble of decisions to produce a reliable final decision.

We further proposed a CNN framework to deal with the joint problem. Although the proposed framework is generic in the sense that any CNN can fit in, we employ a simple CNN architecture with time-frequency image input. The efficiency of this architecture for automatic sleep staging was demonstrated in our previous work \cite{Phan2018c}. To suit the task of joint classification and prediction, we replace the CNN's canonical softmax layer with a \emph{multi-task softmax} layer and introduce the \emph{multi-task loss} function for network training. Without confusion, we will refer to the proposed framework as multi-task framework, joint classification and prediction framework, and one-to-many framework interchangeably throughout this article.

\subsection{Contributions}
The main contributions of this work are as follows.

(i) We formulate automatic sleep staging as a joint classification and prediction problem. The new formulation avoids the shortcomings of the common classification scheme while improving modelling performance.

(ii) A CNN framework is then proposed for the joint problem. To that end, we present and employ a simple and efficient CNN coupled with a multi-task softmax layer and the multi-task loss function to conduct joint classification and prediction.

(iii) We further propose two probabilistic aggregation methods, namely additive and multiplicative voting, to leverage ensemble of decisions available in the proposed framework. 

(iv) Performance-wise, we demonstrate experimentally good performance on two publicly available datasets: Sleep-EDF  \cite{Kemp2000,Goldberger2000} with 20 subjects and  MASS \cite{Oreilly2014}, a large sleep dataset with 200 subjects.

\section{Evaluation Datasets}
\label{sec:datasets}

We used two public datasets: Sleep-EDF Expanded (Sleep-EDF) and Montreal Archive of Sleep Studies (MASS) in this work and conducted analyses under both unimodal (i.e. single-channel EEG) and multimodal conditions (i.e combinations of EEG, EOG, and EMG channels). It should be noted that even though we selected the typical EEG, EOG, and EMG channels in our analyses, the proposed framework, however, can be used straightforwardly to study other signal modalities.

\subsection{Sleep-EDF Expanded (Sleep-EDF)}

Sleep-EDF dataset \cite{Kemp2000, Goldberger2000} consists of two subsets: (1) Sleep Cassette (\emph{SC}) subset consisting of 20 subjects aged 25-34  aiming at studying the age effects on sleep in healthy subjects and (2) Sleep Telemetry (\emph{ST}) subject consisting of 22 Caucasian subjects for study temazepam effects on sleep. We adopted the \emph{SC} subset in this study. PSG recordings, sampled at 100 Hz, of two subsequent day-night periods are available for each subject, except for one subject (subject 13) who has only one-night data. Each 30-second epoch of the recordings was manually labelled by sleep experts according to the R\&K standard \cite{Hobson1969} into one of eight categories \{W, N1, N2, N3, N4, REM, MOVEMENT, UNKNOWN\}. Similar to previous works \cite{Tsinalis2016, Tsinalis2016b, Supratak2017}, N3 and N4 stages were merged into a single stage N3. MOVEMENT and UNKNOWN were excluded. Since full EMG recordings are not available, we only used the Fpz-Cz EEG and the EOG (horizontal) channels in our experiments. Only the in-bed parts (from \emph{lights off} time to \emph{lights on} time) of the recordings were included as recommended in \cite{Imtiaz2014,Imtiaz2015,Tsinalis2016, Tsinalis2016b}.

\subsection{Montreal Archive of Sleep Studies (MASS)}

MASS comprises whole-night recordings from 200 subjects (97 males and 103 females with an age range of 18-76 years). 
These recordings were pooled from different hospital-based sleep laboratories. The available cohort 1 was divided into five subsets of recordings, SS1 - SS5. As stated in the seminal work \cite{Oreilly2014}, heterogeneity between subsets is expected. 
Opposing to the majority of previous works which targeted only one homogeneous subset of the cohort \cite{Dong2017, Supratak2017}, we experimented with all five subsets. Each epoch of the recordings was manually labelled by experts  according to the AASM standard \cite{Iber2007} (SS1 and SS3) and the R\&K standard \cite{Hobson1969}  (SS2, SS4, and SS5). We converted them into five sleep stage \{W, N1, N2, N3, and REM\} as suggested in \cite{Imtiaz2014,Imtiaz2015}. Those recordings with 20-second epochs were converted into 30-second ones by including 5-second segments before and after each epoch. We adopted and studied combinations of the C4-A1 EEG, an average EOG (ROC-LOC), and an average EMG (CHIN1-CHIN2) channels in our experiments. The signals, originally sampled at 256 Hz, were downsampled to 100 Hz. 

\section{Joint Classification and Prediction CNN Framework}
\label{sec:framework}
\subsection{Overview}
\label{ssec:overview}
The proposed framework, with a schematic illustration shown in Figure \ref{fig:overview}, can be described in a stage-wise fashion. The raw signals of a certain epoch index $n$ are first transformed into log-power spectra. The spectra are then preprocessed for frequency smoothing and dimension reduction using frequency-domain filter banks. The resulting channel-specific images are then stacked to form a multi-channel time-frequency image, denoted as $\mathbf{X}_n$. Subsequently, a multi-task CNN is exercised on the multi-channel time-frequency image for joint classification and context prediction. 
The former task is to maximize the conditional probability $P(y_n\,|\, \mathbf{X}_n)$ which characterizes the likelihood of a sleep stage $y_n \in \mathcal{L}=\{1,2,\ldots,Y\}$, where $\mathcal{L}$ denotes the label set of $Y$ sleep stages.
The latter one is to maximize the conditional probabilities $(P(y_{n-\tau}\,|\, \mathbf{X}_n), \ldots, P(y_{n-1}\,|\, \mathbf{X}_n), P(y_{n+1}\,|\, \mathbf{X}_n), \ldots, $ $P(y_{n+\tau}\,|\, \mathbf{X}_n))$ of the neighboring epochs in the output context size of $2\tau + 1$. 
The labels of the epochs in the output context, where $(y_{n-\tau}, \ldots, y_{n}, \ldots, y_{n+\tau})$, can be obtained by probability maximization.

Formally, under this joint classification and prediction formulation, the CNN performs the one-to-many mapping 
\begin{align}
\mathcal{\hat{F}}: \mathbf{X}_n \mapsto (y_{n-\tau},\ldots,y_{n}, \ldots, y_{n+\tau}) \in \mathcal{L}^{2\tau+1}.
\label{eq:joint_class_pred}
\end{align}
Note that the order of the epochs in the neighborhood is encoded by the order of the output labels. This formulation is orthogonal to the common classification one with contextual input of size $2\tau + 1$, in which a network performs the many-to-one mapping
\begin{align}
\mathcal{F}: (\mathbf{X}_{n-\tau}, \ldots, \mathbf{X}_n, \ldots, \mathbf{X}_{n+\tau}) \mapsto y_n \in \mathcal{L}.
\label{eq:class_contextual_input}
\end{align}

Both formulations (\ref{eq:joint_class_pred}) and (\ref{eq:class_contextual_input}) can be interpreted as different extensions of the standard one-to-one classification scheme \cite{Phan2018c, Phan2018d, Andreotti2018}. They will reduce to the standard one when $\tau=0$. However, with our joint classification and prediction formulation, at a certain epoch index $n$ there exists an ensemble of exact $2\tau + 1$ decisions, wherein one classification decision made by itself (i.e. $\mathbf{X}_n$) and $2\tau$ prediction decisions made by its neighbors $(\mathbf{X}_{n-\tau}, \ldots, \mathbf{X}_{n-1}, \mathbf{X}_{n+1}, \ldots, \mathbf{X}_{n+\tau})$. These decisions can be aggregated to form the final decision that is generally better that any individual ones. 

\subsection{Time-Frequency Image Representation}
\label{ssec:representation}

Given a 30-second signal epoch (i.e. EEG, EOG, or EMG), we firstly transform it into a power spectrum using short-time Fourier transform (STFT) with a window size of two seconds and 50\% overlap. Hamming window and 256-point Fast Fourier Transform (FFT) are used. The spectrum is then converted to logarithm scale to produce a log-power spectrum image of size $F \times T$, where $F=129$ and $T=29$.

For frequency smoothing and dimension reduction, the spectrum is filtered by a frequency-domain filter bank. Any frequency-domain filter bank, such as the regular triangular one \cite{Phan2018c}, could serve this purpose. However, it is more favorable to learn the filter bank specifically for the task at hand. Our recent works in \cite{Phan2018c, Phan2018d} demonstrated that a filter bank learned by a DNN in a discriminative fashion is more competent than the regular one in automatic sleep staging. The learned filter bank is expected to emphasize the subbands that are more important for the task and attenuate those less important. Hence, we use the filter bank pretrained with a DNN for preprocessing here. 
One such filter bank with $M=20$ filters is learned for each EEG, EOG, and EMG channel. Filtering the log-power spectrum image reduces its size to $M \times T$. When multiple channels are used, we obtain one such time-frequency image for each channel. For generalization, we denote the time-frequency image as $\mathbf{X} \in \mathbb{R}^{P \times M \times T}$ where $P$ denotes the number of channels. $P=1, 2, 3$ is equivalent to the cases when \{EEG\}, \{EEG, EOG\}, and \{EEG, EOG, EMG\} are employed, respectively.

\subsection{Multi-Task CNN for Joint Classification and Prediction}
\label{ssec:cnn}

\begin{figure} [!t]
	\centering
	\includegraphics[width=0.85\linewidth]{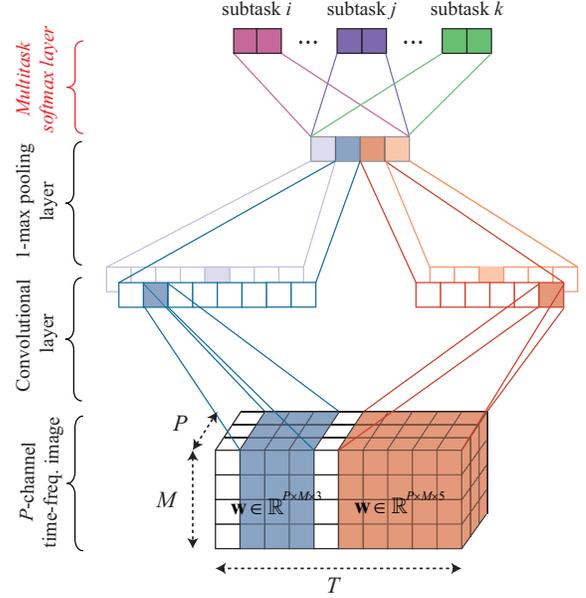}
	\caption{Illustration of the proposed multi-task CNN architecture. The convolution layer of the CNN consists of two filter sets with temporal widths $w=3$ and $w=5$. Each filter set has two individual filters. The colors of the output layer indicate different subtasks jointly modelled by the network.}
	\label{fig:cnn}
\end{figure}

Our recent work \cite{Phan2018c} presented a simple CNN architecture that was shown efficient for sleep staging. We adapt this architecture here by tailoring the last layer, i.e. the multi-task softmax layer, to perform joint classification and prediction. The proposed CNN architecture is illustrated in Figure~\ref{fig:cnn}. Opposing to typical deep CNNs \cite{Stephansen2017,Mikkelsen2018,Tsinalis2016,Supratak2017}, the proposed CNN consists of only three layers: one over-time convolutional layer, one pooling layer, and one \emph{multi-task softmax} layer. This simple architecture has three main characteristics. First, similar to those in \cite{Phan2017,Phan2016,Supratak2017}, its convolutional layer simultaneously accommodates convolutional kernels with varying sizes, and is therefore able to learn features at different resolutions. Second, the exploited \emph{1-max} pooling strategy at the pooling layer is more suitable for capturing the \emph{shift-invariance} property of temporal signals than the common subsampling pooling since a particular feature could occur at any temporal position rather than in a local region of the input signal \cite{Kim2014,Phan2016,Phan2017}. Third, opposing to the canonical softmax, the multi-task softmax layer is adapted to suit the joint classification and prediction. Furthermore, the multi-task loss is introduced for network training. 

Assume that we obtain a training set $\mathcal{S}~=~\left\{\left(\mathbf{X}^{(i)}_{n_i}, (\mathbf{y}^{(i)}_{n_i - \tau}, \ldots, \mathbf{y}^{(i)}_{n_i}, \ldots, \mathbf{y}^{(i)}_{n_i + \tau})\right)\right\}^N_{i=1}$ of size $N$ from the training data. An epoch $i$ is represented by the multi-channel time-frequency image $\mathbf{X}^{(i)}_{n_i} \in \mathbb{R}^{P \times M \times T}$ as described in \ref{ssec:representation} and $n_i$ denotes the corresponding index of the epoch in the original signal. Each epoch $i$ is associated with the sequence of one-hot encoding vectors $(\mathbf{y}^{(i)}_{n_i - \tau}, \ldots, \mathbf{y}^{(i)}_{n_i}, \ldots, \mathbf{y}^{(i)}_{n_i + \tau})$ which represent the sleep stages of the epochs in the context $[n_i-\tau, n_i+\tau]$ of size $2\tau+1$. We use this training set to train the multi-task CNN for joint classification and context prediction.

\begin{figure*} [!t]
	\centering
	\includegraphics[width=0.8\linewidth]{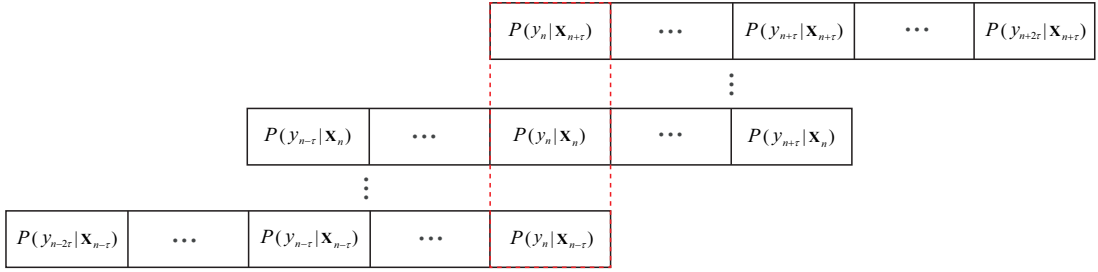}
	\caption{Ensemble of decisions available at the epoch index $n$ made by the epochs $\mathbf{X}_i$ in the neighborhood $[n-\tau, n+\tau]$, i.e. $n-\tau \le i \le n+\tau$.}
	\label{fig:context_smoothing}
\end{figure*}

\subsubsection{Over-Time Convolutional Layer}

Each 3-dimensional filter $\mathbf{w} \in \mathbb{R}^{P \times M \times w}$ of the convolutional layer has the temporal size of $w < T$ while the frequency and channel size entirely cover the frequency and channel dimension of a multi-channel time-frequency image input. The filter is convolved with the input image over time with a stride of $1$. ReLU activation \cite{Nair2010} is then applied to the feature map. 

The CNN is designed to have $R$ filter sets with different temporal widths $w$ to capture features at multiple temporal resolutions. Each filter consists of $Q$ different filters of the same temporal width to allow the CNN to learn multiple complementary features. As a result, the total number of filters is $Q \times R$.

\subsubsection{1-Max Pooling Layer}

We employ 1-max pooling function \cite{Kim2014, Phan2016} on a feature map produced by convolving a filter over an input image to retain the most prominent feature. Pooling all feature maps of $Q \times R$ filters results in a feature vector of size $Q \times R$.

With the over-time convolution layer coupled with the 1-max pooling layer, the CNN functions as a template learning and matching algorithm. The convolutional filters play the role of time-frequency templates that are tuned for the task at hand. Convolving a filter through time can be interpreted as template matching operation, resulting in a feature map that indicates how well the template is matched to different parts of the input image. In turn, 1-max pooling retains a single maximum value, i.e. the maximum matching score, of the feature map as the final feature.

\subsubsection{Multi-Task Softmax Layer}

Opposing to a classification network that typically uses the canonical softmax layer for classification, we propose a multi-task softmax layer to suit joint classification and prediction. The idea is that the network should be penalized for both misclassification and misprediction on a training example. The classification and prediction errors on a training example $i$ is computed as the sum of the cross-entropy errors on the individual subtasks:
\begin{align}
E^{(i)}(\bm{\theta}) = \sum_{n=n_i - \tau}^{n_i +\tau}\mathbf{y}^{(i)}_n\log\left(\hat{\mathbf{y}}^{(i)}_n(\bm{\theta})\right),
\label{eq:multitask_cross_entropy}
\end{align}
where $\bm{\theta}$ and $\hat{\mathbf{y}}$ denote the network parameters and the probability distribution outputted by the CNN, respectively.

The network is trained to minimize the multi-task cross-entropy error over $N$ training samples:
\begin{align}
	E(\bm{\theta}) = -\frac{1}{N}\sum_{i=1}^{N}E^{(i)}(\bm{\theta}) + \frac{\lambda}{2}\|\bm{\theta}\|_2^2.
	\label{eq:multitask_loss}
\end{align}
Here, $\lambda$ denotes the hyper-parameter that trades off the error terms and the $\ell_2$-norm regularization term. For further regularization, \emph{dropout} \cite{Srivastava2014} is also employed. The network training is performed using the \emph{Adam} optimizer \cite{Kingma2015}.

\subsection{Ensemble of Decisions and Aggregation}
\label{ssec:aggregation}

As previously mentioned, one major advantage of the proposed framework is the capacity to produce multiple decisions on a certain epoch even with a single model (the multi-task CNN in this case). Practically, the classification and prediction outputs on a certain epoch may be inconsistent as in ensemble-of-models methods \cite{Hinton2015, Dietterich2000}; aggregation of these multi-view decisions is necessary to derive a more reliable one. 
To that end, we study two probabilistic aggregation schemes: additive and multiplicative voting.

Let $P(y_n\,|\,\mathbf{X}_i)$ denote the estimated probability output on the sleep stage $y_n \in \mathcal{L}$ at the epoch index $n$ given the epoch $\mathbf{X}_i$ in the neighborhood $[n-\tau, n+\tau]$, i.e. $n-\tau \le i \le n+\tau$, as illustrated in Figure \ref{fig:context_smoothing}. The likelihood $P(y_n)$ obtained by additive and multiplicative voting is given by
\begin{align}
P(y_n) = \frac{1}{2\tau+1} \sum_{i=n-\tau}^{n+\tau} P(y_n\,|\,\mathbf{X}_i), \label{eq:additive_smoothing} \\
P(y_n) = \frac{1}{2\tau+1} \prod_{i=n-\tau}^{n+\tau} P(y_n\,|\,\mathbf{X}_i), \label{eq:multiplicative_smoothing}
\end{align}
respectively. Eventually, the predicted label $\hat{y}_n$ is determined by likelihood maximization:
\begin{align}
\hat{y}_n = \argmax_{y_n}P(y_n), \text{~for~} y_n \in \mathcal{L}. \label{eq:likelihood_maximization}
\end{align}

Between the two aggregation schemes, the multiplicative one favors likelihoods of categories with consistent decisions and suppresses likelihoods of those categories with diverged decisions stronger than the additive counterpart \cite{phan2017c}. 

\section{Experiments}
\label{sec:experiment}

We aim at achieving several goals in the conducted experiments. Firstly, we prove empirically the feasibility of predicting labels of the neighboring epochs in the output context concurrently with classifying the current one. Secondly, we demonstrate the advantages of the joint classification and prediction (i.e. many-to-one) formulation over the commonly adopted many-to-one scheme as well as the standard one-to-one classification scheme. Thirdly, we provide performance comparison with various developed baseline systems as well as other deep-learning approaches recently proposed for sleep staging to illustrate the proposed framework's efficiency.

\subsection{Experimental Setup}

For Sleep-EDF, we conducted leave-one-subject-out cross validation. At each iteration, 19 training subjects were further divided into 15 subjects for training and 4 subject for validation. For MASS, we performed 20-fold cross validation on the MASS dataset. At each iteration, 200 subjects were split into training, validation, and test set with 180, 10, and 10 subjects, respectively. The sleep staging performance over 20 folds will be reported for both datasets.

\begin{table}[t!]
	\centering
	\caption{ Parameters of the proposed CNN.}
	\begin{tabular}{|>{\arraybackslash}m{1.0in}|>{\centering\arraybackslash}m{0.75in}|}
		\hline
		{\bf Parameter} & {\bf Value} \parbox{1pt}{\rule{0pt}{2ex+\baselineskip}} \\ [0ex]  	
		\hline
		Filter width $w$ & $\{3, 5, 7\}$ \parbox{1pt}{\rule{0pt}{0.5ex+\baselineskip}} \\ [0ex]
		Number of filters $Q$ & varied \parbox{1pt}{\rule{0pt}{0.5ex+\baselineskip}} \\ [0ex]
		Output context size & 3 \parbox{1pt}{\rule{0pt}{0.5ex+\baselineskip}} \\ [0ex]
		Dropout & $0.2$ \parbox{1pt}{\rule{0pt}{0.5ex+\baselineskip}} \\ [0ex]
		$\lambda$ for regularization & $10^{-3}$ \parbox{1pt}{\rule{0pt}{0.5ex+\baselineskip}} \\ [0ex]
		\hline
	\end{tabular}
	\label{tab:cnn_param}
\end{table}

\begin{table}[t!]
	\caption{The parameters of the deep CNN baseline.}
	\vspace{-0.2cm}
	\begin{center}
		\footnotesize
		\begin{tabular}{|>{\arraybackslash}m{0.25in}|>{\centering\arraybackslash}m{0.3in}|>{\centering\arraybackslash}m{0.25in}|>{\centering\arraybackslash}m{0.45in}|>{\centering\arraybackslash}m{0.5in}|>{\centering\arraybackslash}m{0.01in} @{}m{0pt}@{}}
			\cline{1-5}
			Layer & Size & \#Fmap & Activation & Dropout & \parbox{0pt}{\rule{0pt}{2ex+\baselineskip}} \\ [0ex]  	
			\cline{1-5}
			conv1 & 3 $\times$ 3 & 96 & ReLU & - & \parbox{0pt}{\rule{0pt}{0.5ex+\baselineskip}} \\ [0ex]  	
			pool1 & 2 $\times$ 1 & - & - & 0.2 & \parbox{0pt}{\rule{0pt}{0.5ex+\baselineskip}} \\ [0ex]  	
			conv2 & 3 $\times$ 3 & 96  & ReLU & - & \parbox{0pt}{\rule{0pt}{0.5ex+\baselineskip}} \\ [0ex]  	
			pool2 & 2 $\times$ 2 & - & - & 0.2 &  \parbox{0pt}{\rule{0pt}{0.5ex+\baselineskip}} \\ [0ex]  	
			fc1 & 1024 & - & ReLU & 0.2 & \parbox{0pt}{\rule{0pt}{0.5ex+\baselineskip}} \\ [0ex]  	
			fc2 & 1024 & - & ReLU & 0.2 & \parbox{0pt}{\rule{0pt}{0.5ex+\baselineskip}} \\ [0ex]  	
			\cline{1-5}
		\end{tabular}
	\end{center}
	\label{tab:cnn_baseline}
\end{table}

\subsection{Parameters}
The parameters associated with the proposed CNN are given in Table \ref{tab:cnn_param}. We varied the number of convolutional filters $Q$ of the CNN in the set \{100, 200, 300, 400, 500, 1000\} to investigate its influence. Furthermore, we experimented with the output context size of $3$ (equivalent to $\tau = 1$). Influence of this parameter will be further discussed in Section \ref{sec:discussion}.

The network implementation was based on \emph{Tensorflow} framework \cite{Abadi2016}. Graphic card NVIDIA GTX 1080 Ti was used for network training. The network was trained for 200 epochs with a batch size of 200. The learning rate was set to $10^{-4}$ for the \emph{Adam} optimizer. During training, the network that yielded the best overall accuracy on the validation set was retained for evaluation. Furthermore, we always randomly generated a data batch to have an equal number of samples for all sleep stages to mitigate the class imbalance issue commonly seen in sleep data.

\subsection{Baseline Systems}
\label{ssec:baseline}
To manifest the advantages offered by the proposed frameworks, we constructed two baseline frameworks for comparison:
\begin{itemize}
\item One-to-one: this baseline complies with the standard classification setup, taking a single epoch as input and producing a single decision on its label. 
\item Many-to-one: this baseline conforms to the commonly adopted scheme with contextual input and outputs a single decision on a target epoch. We fixed the contextual input size to 3, i.e. we augmented a target epoch with two nearest neighbors on its left- and right-hand side. 
\end{itemize}

Both baseline frameworks were designed to maintain common experimental settings as those of the proposed one-to-many framework, i.e. the CNN architecture, the learned filter bank, etc. However, it is necessary to use the canonical softmax layer and the standard cross-entropy loss for their classification-only purpose. 

We also developed and repeated the experiments with a typical deep CNN architecture as an alternative to the proposed CNN described in Section \ref{ssec:cnn}. This deep CNN baseline consists of 6 layers (2 convolutional layers, 2 subsampling layers, and 2 fully connected layers) with their parameters characterized in Table \ref{tab:cnn_baseline}.  For simplicity, we refer to our proposed CNN as 1-max CNN to distinguish from the deep CNN baseline. With these experiments, our goal is to show the generalizability of the proposed framework regardless the network base as well as the efficacy of the 1-max CNN in comparison to a typical deep CNN architecture.

\begin{figure*} [!t]
		\centering
		\includegraphics[width=0.9\linewidth]{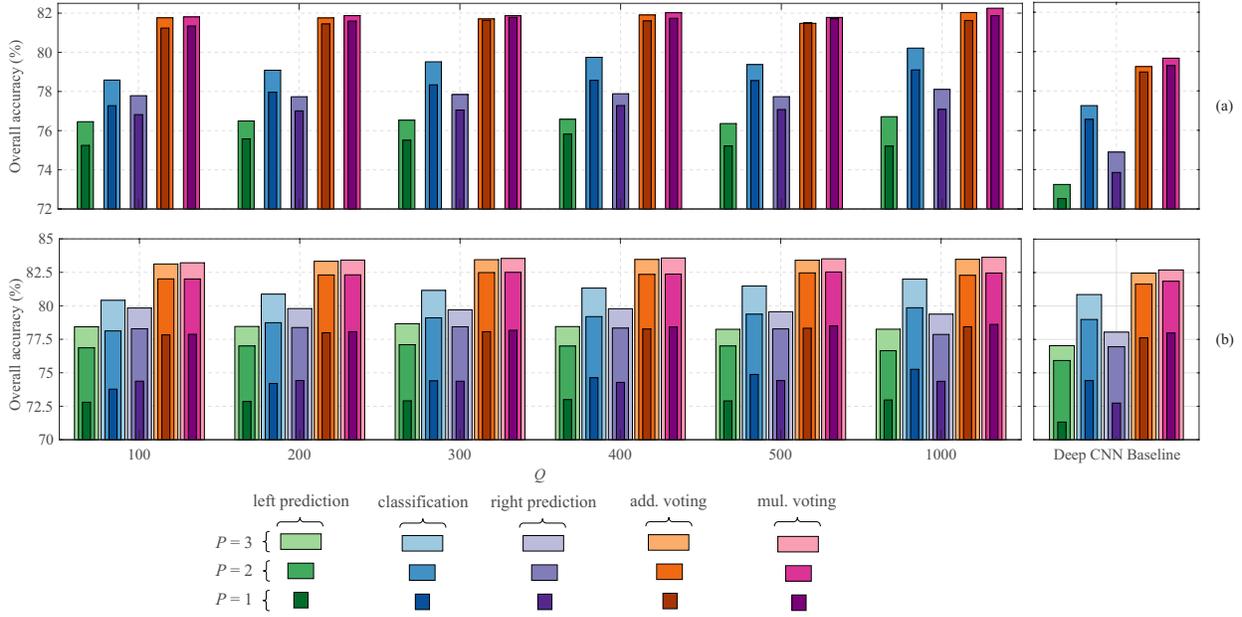}
		\caption{Accuracies of the left prediction subtask, classification subtask, right prediction subtask, multi-task with additive (add.) voting, and multi-task with multiplicative (mul.) voting obtained with an output context size of 3 ($\tau = 1$) and different number of modalities $P$. (a) Sleep-EDF and (b) MASS.}
		\label{fig:predictionvsclassification}
\end{figure*}
\begin{figure} [!t]
		\centering
		\includegraphics[width=1\linewidth]{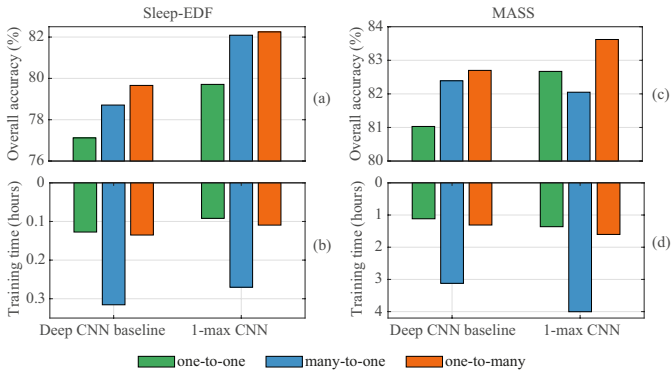}
		\caption{The overall classification accuracy (a)-(c) and the amount of training time (b)-(d) of the proposed framework in comparison with those of the one-to-one, and many-to-one schemes on the first cross-validation fold. We commonly set $Q=1000$ while $P=2$ for Sleep-EDF and  $P=3$ for MASS.}
		\label{fig:inputcontext_vs_outputcontext}
\end{figure}

\subsection{Experimental Results}

\subsubsection{Classification vs prediction accuracy}

In this experiment, we seek to empirically validate the proposed framework by demonstrating the feasibility of context prediction. Since we employed the output context size of $3$, without confusion, let us refer to the network's subtasks as \emph{classification}, \emph{left prediction}, and \emph{right prediction}, which correspond to decisions on the input epoch, its left neighbor, and its right neighbor.

We show in Figure \ref{fig:predictionvsclassification} the accuracy rates of classification, left prediction, and right prediction subtasks obtained by the 1-max CNN (with varying number of convolutional filters $Q$) and the deep CNN baseline with the different number of input modalities $P$. Unlike the classification subtask, the CNNs do not have access to the signal information of the left and right neighboring epochs. As a result, inference for their labels relies solely on their dependency with the input epoch. It can be expected that the accuracy rates of the left and right prediction subtasks are lower than that of the classification subtask in most of the cases. Nevertheless, overall both CNNs maintain a good accuracy level in prediction relative to the classification accuracy, especially in multimodal cases (e.g. $P=2$ for Sleep-EDF and $P=3$ for MASS).
More specifically, averaging over all $Q$ and $P$, the left and right prediction accuracies of the 1-max CNN are only $2.9\%$ and $1.4\%$ lower than the classification accuracy on Sleep-EDF whereas the respective gaps of $2.2\%$ and $1.3\%$ are seen in MASS. Similar patterns can also be seen with the deep CNN baseline with the graceful degradation of $4.0\%$ and $2.5\%$ in Sleep-EDF and $3.3\%$ and $2.2\%$ in MASS correspondingly.
These results strengthen the assumption about the dependency between neighboring PSG epochs and consolidate the feasibility of joint classification and prediction modelling.

\subsubsection{Advantages of the joint classification and prediction}

Figure \ref{fig:predictionvsclassification} also highlights the performance improvements obtained by the joint classification and prediction framework after the aggregation step in comparison to individual subtasks. 
Averaging over all $P$ and $Q$, the 1-max CNN with additive voting leads to $2.8\%$ and $4.5\%$ absolute accuracy gains over the classification subtask's accuracy on Sleep-EDF and MASS, respectively. The gains yielded by the multiplicative voting are even better, reaching $3.0\%$ and $4.7\%$, respectively. Accordingly, the deep CNN baseline produces $2.2\%$ and $2.5\%$ absolute gains with additive voting and $2.6\%$ and $2.8\%$ with multiplicative voting on the two datasets. Between two voting schemes, the performance gain of the multiplicative one is slightly better than that of the additive counterpart with a difference around $0.2 - 0.3\%$ on both Sleep-EDF and MASS. 

To demonstrate the advantages of the proposed framework over the common classification schemes, we further compare its performance and computational complexity with the one-to-one and many-to-one baseline schemes described in Section \ref{ssec:baseline}. 
For simplicity, we utilized all available modalities (i.e. $P=3$) in this experiment and made use of multiplicative-voting aggregation in the proposed framework. Additionally, we set the number of convolutional filters $Q=1000$ when the 1-max CNN was employed.

Figure \ref{fig:inputcontext_vs_outputcontext} depicts the overall accuracy obtained by the three frameworks and their computational complexity in terms of the training time. Note that we only included the training time of the first cross-validation fold as a representative here and the training time was expected to scale linearly with the amount of training data. Four important points should be noticed from the figure. Firstly, contextual input does not always help as the many-to-one baseline with the 1-max CNN experiences a performance drop of $0.6\%$ absolute compared to the one-to-one on MASS compared to the one-to-one on MASS although it improves accuracy rates in other cases. Secondly, the proposed one-to-many framework consistently outperforms its counterparts. Adopting the 1-max CNN as the base, our framework outperforms the one-to-one and many-to-one opponents with 
$2.5\%$ and $0.2\%$ absolute in Sleep-EDF and $1.0\%$ and $1.6\%$ absolute in MASS, respectively. Similar gains of $2.5\%$ and $1.0\%$ in Sleep-EDF; $1.7\%$ and $0.3\%$ in MASS are achieved when the deep CNN baseline is used.
Thirdly, between the network bases, the 1-max CNN surpasses the deep CNN baseline with an improvement of $2.6\%$ absolute in Sleep-EDF and $0.9\%$ absolute in MASS although its architecture is much simpler. 
Fourthly, concerning the computational complexity, three times larger input of the many-to-one baseline roughly triples the training time compared to that of the one-to-one. For instance, 4.0 hours versus 1.36 hours in MASS can be seen with the 1-max CNN. Differently, with the training time of $1.6$ hours. Using the same network, the proposed framework only increases computing time by as small as $0.2$ hours. The training time of the deep CNN baseline also exposes similar patterns. 

\setlength\tabcolsep{2.5pt} 
\begin{table*}[!t]
	\caption{Performance comparison of different systems developed in this work. We marked in bold the figures where the combination of the one-to-many framework and 1-max CNN outperforms all other opponents.}
	\vspace{-0.2cm}
	\footnotesize
	\begin{center}
		\begin{tabu}{|>{\arraybackslash}m{0.1in}|>{\arraybackslash}m{1.6in}|>{\centering\arraybackslash}m{0.25in}|>{\centering\arraybackslash}m{0.25in}|>{\centering\arraybackslash}m{0.25in}|>{\centering\arraybackslash}m{0.25in}|>{\centering\arraybackslash}m{0.25in}|>{\centering\arraybackslash}m{0.25in}|>{\centering\arraybackslash}m{0.25in}|>{\centering\arraybackslash}m{0.25in}|>{\centering\arraybackslash}m{0.25in}|>{\centering\arraybackslash}m{0.25in}|>{\centering\arraybackslash}m{0.25in}|>{\centering\arraybackslash}m{0.25in}|>{\centering\arraybackslash}m{0.25in}|>{\centering\arraybackslash}m{0.25in}|>{\centering\arraybackslash}m{0.3in}|>{\centering\arraybackslash}m{0in} @{}m{0pt}@{}}
			\cline{3-17}
			\multicolumn{2}{c|}{} & \multicolumn{5}{c|}{$P=1$ (EEG only)} & \multicolumn{5}{c|}{$P=2$ (EEG + EOG)} & \multicolumn{5}{c|}{$P=3$ (EEG + EOG + EMG)} & \parbox{0pt}{\rule{0pt}{2ex+\baselineskip}} \\ [0ex]  	
			\cline{3-17}
			\multicolumn{2}{c|}{} & Acc. & $\kappa$ & MF1 & Sens. & Spec. &  Acc. & $\kappa$ & MF1 & Sens. & Spec. & Acc. & $\kappa$ &   MF1 & Sens. & Spec. &  \parbox{0pt}{\rule{0pt}{2ex+\baselineskip}} \\ [0ex]  	
			\cline{1-17}
			\multirow{6}{*}{\begin{sideways}{Sleep-EDF}\end{sideways}} & \emph{\bf One-to-many + 1-max CNN} & $\bm{81.9}$ & $\bm{0.74}$ & $\bm{73.8}$ & $\bm{73.9}$ & $\bm{95.0}$ & $\bm{82.3}$ & $\bm{0.75}$ & $74.7$ & $74.3$ & $\bm{95.1}$ & \multicolumn{4}{c}{} & \parbox{0pt}{\rule{0pt}{0.5ex+\baselineskip}} \\ [0ex]  	
			& \emph{One-to-one + 1-max CNN} &  $79.8$ & $0.72$ & $72.0$ & $72.4$ & $94.6$ & $79.7$ & $0.72$ & $72.2$ & $72.8$ & $94.6$ & \multicolumn{4}{c}{} & \parbox{0pt}{\rule{0pt}{0.5ex+\baselineskip}} \\ [0ex]  	
			& \emph{Many-to-one + 1-max CNN} & $80.9$ & $0.73$ & $73.6$ & $74.2$ & $94.9$ & $82.1$ & $0.75$ & $75.4$ & $75.4$ & $95.1$ & \multicolumn{4}{c}{} & \parbox{0pt}{\rule{0pt}{0.5ex+\baselineskip}} \\ [0ex]  	
			& \emph{One-to-many + deep CNN baseline} & $79.3$ & $0.71$ & $69.7$ & $70.2$ & $94.2$ & $79.7$ & $0.71$ & $71.2$ & $70.9$ & $94.3$ & \multicolumn{4}{c}{} & \parbox{0pt}{\rule{0pt}{0.5ex+\baselineskip}} \\ [0ex]  	
			& \emph{One-to-one + deep CNN baseline} & $76.7$ & $0.67$ & $67.6$ & $68.6$ & $93.7$ & $77.1$ & $0.68$ & $69.3$ & $69.8$ & $93.8$ & \multicolumn{4}{c}{} & \parbox{0pt}{\rule{0pt}{0.5ex+\baselineskip}} \\ [0ex]  	
			& \emph{Many-to-one + deep CNN baseline} & $78.3$ & $0.69$ & $70.7$ & $71.1$ & $94.1$ & $78.7$ & $0.70$ & $71.8$ &  $72.4$ & $94.2$ & \multicolumn{4}{c}{} & \parbox{0pt}{\rule{0pt}{0.5ex+\baselineskip}} \\ [0ex]  	
			
			\cline{1-17}
			\cline{1-17}
			\multirow{6}{*}{\begin{sideways}{MASS}\end{sideways}} & \emph{\bf One-to-many + 1-max CNN} &  $\bm{78.6}$ & $\bm{0.70}$ & $\bm{70.6}$ & $71.2$ & $\bm{94.1}$ & $\bm{82.5}$ & $\bm{0.75}$ & $\bm{76.1}$ & $75.8$ & $\bm{95.0}$ & $\bm{83.6}$ & $\bm{0.77}$ & $77.9$ & $77.4$ & $\bm{95.3}$ \parbox{0pt}{\rule{0pt}{0.5ex+\baselineskip}} \\ [0ex]  	
			& \emph{One-to-one + 1-max CNN} &  $75.9$ & $0.67$ & $69.6$ & $71.1$ & $93.7$ & $80.7$ & $0.73$ & $74.9$ & $75.5$ & $94.8$ & $82.7$ & $0.75$ & $77.6$ & $77.8$ & $95.1$ \parbox{0pt}{\rule{0pt}{0.5ex+\baselineskip}} \\ [0ex]  	
			& \emph{Many-to-one + 1-max CNN} & $76.3$ & $0.67$ & $69.8$ & $71.3$ & $93.8$ & $80.9$ & $0.73$ & $75.1$ & $75.5$ & $94.8$ & $82.1$ & $0.75$ & $76.6$ & $76.9$ & $95.0$ \parbox{0pt}{\rule{0pt}{0.5ex+\baselineskip}} \\ [0ex]  	
			
			& \emph{One-to-many + deep CNN baseline} &  $78.0$ & $0.69$ & $69.8$ & $70.1$ & $93.8$ & $81.9$ & $0.74$ & $75.2$ & $74.7$ & $94.8$ & $82.7$ & $0.75$ & $76.9$ & $76.3$ & $95.0$ \parbox{0pt}{\rule{0pt}{0.5ex+\baselineskip}} \\ [0ex]  	
			& \emph{One-to-one + deep CNN baseline} & $74.5$ & $0.65$ & $68.4$ & $70.0$ & $93.4$ & $79.2$ & $0.71$ & $73.5$ & $74.3$ & $94.4$ & $81.0$ & $0.73$ & $76.4$ & $77.4$ & $94.9$ \parbox{0pt}{\rule{0pt}{0.5ex+\baselineskip}} \\ [0ex]  	
			& \emph{Many-to-one + deep CNN baseline} & $77.4$ & $0.68$ & $71.6$ & $72.8$ & $94.0$ & $81.2$ & $0.73$ & $76.0$ & $76.4$ & $94.8$ & $82.4$ & $0.75$ & $78.2$ & $78.9$ & $95.2$ \parbox{0pt}{\rule{0pt}{0.5ex+\baselineskip}} \\ [0ex]  	
			\cline{1-17}
		\end{tabu}
	\end{center}
	\label{tab:performance}
\end{table*}

\setlength\tabcolsep{2.5pt} 
\begin{table*}[t!]
		\caption{Performances of the proposed method compared to previous methods on the Sleep-EDF dataset. Notice the large variation in the accuracy rate due to the differences in experimental setup. Top accuracy rates, such as in Aboalayon \emph{et al.} \cite{Aboalayon2016}, Alickovic \& Subasi \cite{Alickovic2018}, and Dimitriadis \emph{et al.} \cite{Dimitriadis2018}, are likely biased by nonindependent testing and usage of entire recordings rather than only in-bed data (cf. \ref{sssec:performance_comparison} for further detail).}
		\vspace{-0.2cm}
		\scriptsize
		\begin{center}
			\begin{tabular}{|>{\arraybackslash}m{1.2in}|>{\arraybackslash}m{0.9in}|>{\arraybackslash}m{0.8in}|>{\arraybackslash}m{0.5in}|>{\centering\arraybackslash}m{0.6in}|>{\centering\arraybackslash}m{0.5in}|>{\centering\arraybackslash}m{0.4in}|>{\centering\arraybackslash}m{0.4in}|>{\centering\arraybackslash}m{0in} @{}m{0pt}@{}}
				\cline{2-8}
				\multicolumn{1}{c|}{} & Method & Input channel & Feature type & Subjects & Independent testing & In-bed data only & Overall accuracy &  \parbox{0pt}{\rule{0pt}{2ex+\baselineskip}} \\ [0ex]  	
				
				\cline{1-8}		
				\bf This work & Multitask 1-max CNN &  Fpz-Cz + hor. EOG & learned &  20 SC & yes & yes & $82.3$ & \parbox{0pt}{\rule{0pt}{0.5ex+\baselineskip}} \\ [0ex]  	
				
				\bf This work & Multitask 1-max CNN &  Fpz-Cz & learned &  20 SC & yes & yes & $81.9$ & \parbox{0pt}{\rule{0pt}{0.5ex+\baselineskip}} \\ [0ex]  	
				
				Phan \emph{et al.} \cite{Phan2018c} & 1-max CNN &  Fpz-Cz & learned &  20 SC & yes & yes & $79.8$ & \parbox{0pt}{\rule{0pt}{0.5ex+\baselineskip}} \\ [0ex]  	
				
				Phan \emph{et al.} \cite{Phan2018d} & Attentional RNN &  Fpz-Cz & learned &  20 SC & yes & yes & $79.1$ & \parbox{0pt}{\rule{0pt}{0.5ex+\baselineskip}} \\ [0ex]  	
				
				Andreotti \emph{et al.} \cite{Andreotti2018} & ResNet &  Fpz-Cz + hor. EOG & learned &  20 SC & yes & yes & $76.8$ & \parbox{0pt}{\rule{0pt}{0.5ex+\baselineskip}} \\ [0ex]  	
				
				Tsinalis \emph{et al.} \cite{Tsinalis2016b} & Deep auto-encoder &  Fpz-Cz & hand-crafted &  20 SC & yes & yes & $78.9$ & \parbox{0pt}{\rule{0pt}{0.5ex+\baselineskip}} \\ [0ex]  	
				
				Tsinalis \emph{et al.} \cite{Tsinalis2016} & Deep CNN & Fpz-Cz & learned &  20 SC & yes & yes & $74.8$ &\parbox{0pt}{\rule{0pt}{0.5ex+\baselineskip}} \\ [0ex]  	
				
				Supratak \emph{et al.} \cite{Supratak2017} & Deep CNN + RNN &  Fpz-Cz & learned &  20 SC & yes & no & $82.0$ & \parbox{0pt}{\rule{0pt}{0.5ex+\baselineskip}} \\ [0ex]  	
				
				Alickovic \& Subasi \cite{Alickovic2018} &  Ensemble SVM & Pz-Oz & hand-crafted & 10 SC + 10 ST & yes & no & $91.1$ & \parbox{0pt}{\rule{0pt}{0.5ex+\baselineskip}} \\ [0ex]  	
				
				Sanders \emph{et al.} \cite{Sanders2014} & Decision trees & Fpz-Cz & hand-crafted & 10 ST & yes & no & $75.0$ & & \parbox{0pt}{\rule{0pt}{0.5ex+\baselineskip}} \\ [0ex]  	
				
				Dimitriadis \emph{et al.} \cite{Dimitriadis2018} & $k$-NN & Fpz-Cz & hand-crafted & 20 SC & yes &  no & $94.4$ &  \parbox{0pt}{\rule{0pt}{0.5ex+\baselineskip}} \\ [0ex]  	
				
				Mikkelsen \& De Vos \cite{Mikkelsen2018} &  Deep CNN & Fpz-Cz + hor. EOG & learned & 20 SC & no  & yes & $84.0$ &\parbox{0pt}{\rule{0pt}{0.5ex+\baselineskip}} \\ [0ex]  	
				
				Imtiaz \emph{et al.} \cite{Imtiaz2015b} &  Ensemble SVM & Fpz-Cz + Pz-Oz & hand-crafted & 20 SC + 22 ST & no &  yes & $78.9$  & \parbox{0pt}{\rule{0pt}{0.5ex+\baselineskip}} \\ [0ex]  	
				
				Munk \emph{et al.} \cite{Munk2018} & GMM & Pz-Oz & hand-crafted & 19 SC & no  & no & $73.2$ & \parbox{0pt}{\rule{0pt}{0.5ex+\baselineskip}} \\ [0ex]  	
				
				Rodr\'{i}guez-Sotelo \emph{et al.} \cite{Rodriguez-Sotelo2014} & $k$-NN & Fpz-Cz + Pz-Oz & hand-crafted & 20 SC & no &  no & $80.0$ &  \parbox{0pt}{\rule{0pt}{0.5ex+\baselineskip}} \\ [0ex]  	
				
				Aboalayon \emph{et al.} \cite{Aboalayon2016} & Decision trees & Fpz-Cz + Pz-Oz & hand-crafted & 20 SC & no &  no & $93.1$ &  \parbox{0pt}{\rule{0pt}{0.5ex+\baselineskip}} \\ [0ex]  	
				\cline{1-8}
			\end{tabular}
		\end{center}
		\label{tab:performance_comparison_EDF}
\end{table*}

\setlength\tabcolsep{2.5pt} 
\begin{table*}[t!]
		\caption{Performances of the proposed method compared to previous methods on the MASS dataset.}
		\vspace{-0.2cm}
		\scriptsize
		\begin{center}
			\begin{tabular}{|>{\arraybackslash}m{0.9in}|>{\arraybackslash}m{0.9in}|>{\arraybackslash}m{1.5in}|>{\arraybackslash}m{0.5in}|>{\centering\arraybackslash}m{0.6in}|>{\centering\arraybackslash}m{0.5in}|>{\centering\arraybackslash}m{0.4in}|>{\centering\arraybackslash}m{0in} @{}m{0pt}@{}}
				\cline{2-7}
				\multicolumn{1}{c|}{} & Method & Input channel & Feature type & Subjects & Independent testing & Overall accuracy &  \parbox{0pt}{\rule{0pt}{2ex+\baselineskip}} \\ [0ex]  	
				
				\cline{1-7}		
				
				\bf This work & Multitask 1-max CNN &  C4-A1 + ROC-LOC + CHIN1-CHIN2 & learned &  200 & yes  & $83.6$ & \parbox{0pt}{\rule{0pt}{0.5ex+\baselineskip}} \\ [0ex]  	
				
				\bf This work & Multitask 1-max CNN &  C4-A1 + ROC-LOC & learned &  200 & yes & $82.5$ & \parbox{0pt}{\rule{0pt}{0.5ex+\baselineskip}} \\ [0ex]  	
				
				\bf This work & Multitask 1-max CNN &  C4-A1 & learned & 200  & yes &  $78.6$ & \parbox{0pt}{\rule{0pt}{0.5ex+\baselineskip}} \\ [0ex]  	
				
				\emph{Chambon et al.}\textsuperscript{2} \cite{Chambon2018} & Deep CNN &  C4-A1 + ROC-LOC + CHIN1-CHIN2 & learned & 200 & yes &  $79.9$ & \parbox{0pt}{\rule{0pt}{0.5ex+\baselineskip}} \\ [0ex]  	
								
				\emph{DeepSleepNet1}\textsuperscript{2} \cite{Supratak2017} & Deep CNN &  C4-A1 + ROC-LOC + CHIN1-CHIN2 & learned & 200 & yes &  $80.7$ & \parbox{0pt}{\rule{0pt}{0.5ex+\baselineskip}} \\ [0ex]  	
				
				\emph{Tsinalis et al.}\textsuperscript{2} \cite{Tsinalis2016} & Deep CNN &  C4-A1 + ROC-LOC + CHIN1-CHIN2 & learned & 200 & yes & $77.9$ & \parbox{0pt}{\rule{0pt}{0.5ex+\baselineskip}} \\ [0ex]  	
				
				Andreotti \emph{et al.} \cite{Andreotti2018} & ResNet &  C4-A1 + ROC-LOC + CHIN1-CHIN2 & learned & 200 & yes & $79.4$ & \parbox{0pt}{\rule{0pt}{0.5ex+\baselineskip}} \\ [0ex]  	
				
				Chambon \emph{et al.} \cite{Chambon2018} & Deep CNN & 6 EEG + 2 EOG + 3 EMG & learned &  61 (SS3 only) & yes &  $83.0$ & \parbox{0pt}{\rule{0pt}{0.5ex+\baselineskip}} \\ [0ex]  	
				
				Supratak \emph{et al.} \cite{Supratak2017} & Deep CNN &  F4-EOG (left) & learned &  62 (SS3 only) & yes & $81.5$ & \parbox{0pt}{\rule{0pt}{0.5ex+\baselineskip}} \\ [0ex]  	
				
				Dong \emph{et al.} \cite{Dong2017} & DNN &  F4-EOG (left) & learned &  62 (SS3 only) & yes &  $81.4$ & \parbox{0pt}{\rule{0pt}{0.5ex+\baselineskip}} \\ [0ex]  	
				
				Dong \emph{et al.} \cite{Dong2017} & Random Forests &  F4-EOG (left) &  hand-crafted &  62 (SS3 only) & yes &  $81.7$ & \parbox{0pt}{\rule{0pt}{0.5ex+\baselineskip}} \\ [0ex]  	
				
				Dong \emph{et al.} \cite{Dong2017} & SVM &  F4-EOG (left) & hand-crafted &  62 (SS3 only) & yes &  $79.7$ & \parbox{0pt}{\rule{0pt}{0.5ex+\baselineskip}} \\ [0ex]  	

				\cline{1-7}		
			\end{tabular}
		\end{center}
		\label{tab:performance_comparison_MASS}
\end{table*}

\subsubsection{Performance comparison}
\label{sssec:performance_comparison}
Table \ref{tab:performance} provides a comprehensive performance comparison on the experimental dataset using different metrics, including overall accuracy, kappa index $\kappa$, average specificity, average sensitivity, and average macro F1-score (MF1). The comparison covers all combinations of different frameworks (i.e. the proposed and the baselines) and network bases (i.e. the proposed 1-max CNN and the deep CNN baseline). \blfootnote{\footnotesize \textsuperscript{2}Our implementation. Source code is also available at \url{http://github.com/pquochuy/MultitaskSleepNet}}As can be seen, the proposed one-to-many framework powered by the 1-max CNN ({\bf one-to-many + 1-max CNN}) outperforms other combinatorial systems presented in this work on both datasets and over different combinations of modalities. There are occasional exceptions where using the 1-max CNN in the baseline frameworks yields marginally better average MF1 and Sensitivity than {\bf one-to-many + 1-max CNN}, such as on MASS with $P=3$; however, {\bf one-to-many + 1-max CNN} remains optimal on other metrics.

To see an overall picture, in Tables \ref{tab:performance_comparison_EDF} and \ref{tab:performance_comparison_MASS} we relate the proposed method's accuracy to those reported by previous works on the two datasets. With this comprehensive comparison, we also aim at providing a benchmark for future work.
	
As can be seen from Table \ref{tab:performance_comparison_EDF}, the results on Sleep-EDF vary noticeably due to the lack of standardization in experimental setup. We observe two factors that greatly affects performance on this dataset: (1) independent/dependent testing and (2) whether or not using only \emph{in-bed} parts of the recordings as recommended in \cite{Imtiaz2015,Imtiaz2015b,Tsinalis2016, Tsinalis2016b}. Dependent testing happens when data of a test subject is also involved in training, such as in Aboalayon \emph{et al.} \cite{Aboalayon2016}, and biases the evaluation results. In addition to in-bed parts (i.e. from \emph{lights off} time to \emph{lights on} time \cite{Imtiaz2015}), many previous studies also included other parts, such as in Supratak \emph{et al.} \cite{Supratak2017}, or even entire recordings, such as in Dimitriadis \emph{et al.} \cite{Dimitriadis2018} and in Alickovic \& Subasi \cite{Alickovic2018}, into their experiments. These add-on data, which are mainly \emph{Wake} epochs, often boost the performance as \emph{Wake}, in general, is easier to be recognized than other sleep stages. Therefore, the performance comparison is improper unless two methods use a similar experimental setup. With respect to this, the proposed method outperforms other competitors that commonly used independent testing and in-bed data only. It should be noted that these results do not cover a large body of studies on the early version of Sleep-EDF dataset \cite{Kemp2000,Goldberger2000} which consists of only 8 PSG recordings.

A few recent attempts has evaluated automatic sleep staging on a subset \cite{Supratak2017,Dong2017,Chambon2018} rather than the entire 200 subjects of the MASS dataset. The discrepancy in data makes a direct comparison between their results and ours inappropriate. To avoid possible mismatch in experimental setup, we re-implemented DeepSleepNet \cite{Supratak2017} and the deep CNN architecture proposed by Chambon \emph{et al.} \cite{Chambon2018}, both of which recently reported the state-of-the-art results on the MASS subset SS3, for a compatible comparison. Note that we experimented with DeepSleepNet1 (CNN) in \cite{Supratak2017} here, and will leave DeepSleepNet2 (CNN combined with RNN for long-term context modelling) for future work. In addition, we also implemented the deep CNN proposed by Tsinalis \emph{et al.} \cite{Tsinalis2016} which demonstrated good performance on Sleep-EDF.
While our developed baselines (cf. Table \ref{tab:performance}) are more efficient than these networks under the common experimental setup used in this work, the improvements by the proposed multitask 1-max CNN are most prominent, as can be seen from Table \ref{tab:performance_comparison_MASS}. More specifically, compared to the best opponent, DeepSleepNet \cite{Supratak2017}, a margin of $2.9\%$ on overall accuracy is obtained when all three adopted channels ($P=3$) were used.
	
\begin{figure*} [!t]
	\centering
	\includegraphics[width=0.85\linewidth]{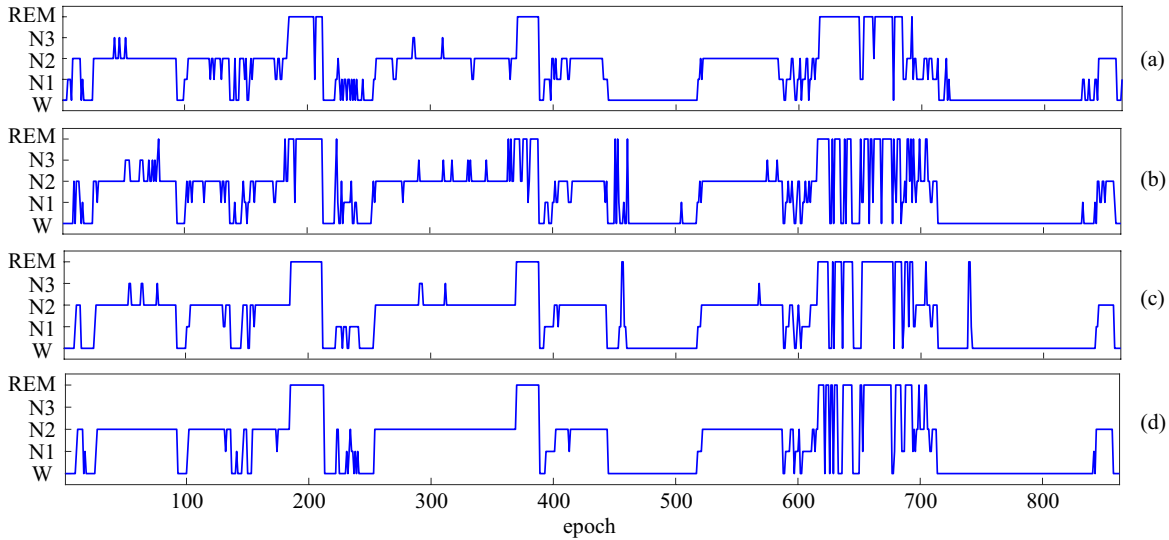}
	\caption{Hypnogram of one subject of the MASS dataset (subject 22 of the subset SS1 \cite{Oreilly2014}): (a) ground-truth, (b) the one-to-one baseline framework's output, (c) the many-to-one baseline framework's output, (d) the proposed one-to-many framework's output. 1-max CNN was commonly used with $Q=100$ and $P=3$.}
	\label{fig:hypnogram_mass}
\end{figure*}

For completeness, we show in Table \ref{tab:confusion_matrix} the confusion matrices and class-wise performance in terms of sensitivity and selectivity \cite{Imtiaz2014} obtained by the proposed one-to-many framework with the 1-max CNN base. Particularly, one may notice modest performance on N1 stage, which has been proven challenging to be correctly recognized \cite{Phan2018c, Phan2018d, Supratak2017,Tsinalis2016} due to its similarities with other stages and its infrequency. Possibilities for improvement would be to over-sample the under-present class during training and to explore weighting schemes for a network's loss \cite{phan2018e,Koch2018b} so that the network is penalized stronger if making errors on this infrequent class than other ones. We further provide alignment of ground-truth and system-output hypnograms for one subject of the MASS dataset in Figure~\ref{fig:hypnogram_mass}.

\setlength\tabcolsep{2.5pt} 
\begin{table}[t!]
		\caption{Confusion matrices and class-wise performance (sensitivity and selectivity) obtained by the proposed one-to-many framework with the 1-max CNN base.}
		\footnotesize
		\vspace{-0.2cm}
		\begin{center}
			\begin{tabular}{|>{\arraybackslash}m{0.25in}|>{\centering\arraybackslash}m{0.25in}|>{\centering\arraybackslash}m{0.3in}|>{\centering\arraybackslash}m{0.3in}|>{\centering\arraybackslash}m{0.3in}|>{\centering\arraybackslash}m{0.3in}|>{\centering\arraybackslash}m{0.3in}|>{\centering\arraybackslash}m{0.25in}|>{\centering\arraybackslash}m{0.25in}|>{\centering\arraybackslash}m{0in} @{}m{0pt}@{}}
				\cline{3-9}
				\multicolumn{2}{c|}{} & \multicolumn{5}{c|}{Groundtruth} & \multirow{2}{*}{\makecell{Sen. \\ (\%)}}& \multirow{2}{*}{\makecell{Sel. \\ (\%)}} \parbox{0pt}{\rule{0pt}{0.5ex+\baselineskip}} \\ [0ex]  	
				\cline{3-7}
				\multicolumn{2}{c|}{} & W & N1 & N2 & N3 & REM &  & & \parbox{0pt}{\rule{0pt}{0.5ex+\baselineskip}} \\ [0ex]  	
				\cline{1-9}
				\multirow{5}{*}{\begin{sideways}{\makecell{Sleep-EDF\\Output}}\end{sideways}}& W & 3403	& 322	& 230	& 32	& 522 & 75.5 & 79.3 &\parbox{0pt}{\rule{0pt}{0ex+\baselineskip}} \\ [0ex]  	
				& N1 & 441	& 880	& 725	& 9	& 707 & 31.9 & 55.7 &\parbox{0pt}{\rule{0pt}{0ex+\baselineskip}} \\ [0ex]  	
				& N2& 230	& 263	& 15263	& 795	& 1026 & 86.8 & 88.1 &\parbox{0pt}{\rule{0pt}{0ex+\baselineskip}} \\ [0ex]  	
				& N3 & 65	& 0	& 658	& 4850	& 18 & 86.7& 85.3 &\parbox{0pt}{\rule{0pt}{0ex+\baselineskip}} \\ [0ex]  	
				& REM &  154	& 114 &	457	& 3	& 6983 & 90.6 & 75.4 &\parbox{0pt}{\rule{0pt}{0ex+\baselineskip}} \\ [0ex]  	
				\cline{1-9}
				\multirow{5}{*}{\begin{sideways}{\makecell{MASS \\ Output}}\end{sideways}}& W & 26261 &	2148 &	1450 &	72 &	1112 &	84.6 &	86.3 &\parbox{0pt}{\rule{0pt}{0ex+\baselineskip}} \\ [0ex]  	
				& N1 & 2924 &	7948 &	5498 &	22	& 2965	& 41.1 &	55.2 &\parbox{0pt}{\rule{0pt}{0ex+\baselineskip}} \\ [0ex]  	
				& N2& 759	& 3429	& 95486	& 4849	& 3395	& 88.5 & 86.9 &\parbox{0pt}{\rule{0pt}{0ex+\baselineskip}} \\ [0ex]  	
				& N3 & 30	& 13 &	6098 &	24223	& 18	& 79.7 &	83.0 &\parbox{0pt}{\rule{0pt}{0ex+\baselineskip}} \\ [0ex]  	
				& REM &  466	& 872	& 1353	& 6	& 37473	& 93.3 &	83.3 &\parbox{0pt}{\rule{0pt}{0ex+\baselineskip}} \\ [0ex]  	
				\cline{1-9}
			\end{tabular}
		\end{center}
		\label{tab:confusion_matrix}
\end{table}

\section{Discussion}
\label{sec:discussion}

In this section, we investigate the causes of the proposed framework's performance improvement over the baseline ones. Furthermore, the proposed framework encompasses several influential factors, such as the number of convolutional filters $Q$ of the 1-max CNN, the number of input modalities $P$, and  the output context size.  We will discuss and elucidate their effects on the framework's performance. 
The multitask framework will also be contrasted against an equivalent ensemble method to shed light on their similar behaviour.

\subsection{Investigating the Causes of Improvement}

To accomplish this goal, we divided the dataset into a \emph{non-transition} and \emph{transition} set and explored how different frameworks perform on them. Considering MASS for this investigation, the former set is the major one ($83.4\%$ epochs in total) consisting of epochs with the same label as their left and right neighbors. The latter, which is the minor set ($16.6\%$ epochs in total), comprises those epochs at stage transitions, i.e. their labels differ from those of their left/right neighbors or both.

The overall accuracy on these sets are shown in Table \ref{tab:performance_transition_nontransition}. On one hand, the downgrading accuracy on the transition set reflects the fact that manual labelling of sleep stages if of low accuracy near stage transitions \cite{Rosenberg2014}. Since a 30-second epoch likely contains the signal information of two transitioning stages while only one label is assigned to such an epoch, up to half of the epoch may not match the assigned label. More often than not, the labels assigned to these epochs are subjective to the scorer. The accuracy of the one-to-one baseline framework on this small subset, which is above the chance level, is likely due to the bias towards the scorer's subjectivity. The chance-level accuracy of the many-to-one and one-to-many frameworks, on the other hand, can be explained by the fact that taking into account the left and right neighboring epochs has balanced the contribution of the two transitioning stages.

Disregarding the ambiguous transition set, the cause of performance improvement turns out to be depending upon the accuracy on the major non-transition set. As can be seen, the proposed framework outperforms the other two with a gap of $2.7\%$ and $1.3\%$ on this set, respectively. Further investigation on this set reveals a substantial level of label agreement between the proposed framework and the one-to-one baseline, up to $91.0\%$. However, for the remaining $9.0\%$ epochs on which their labels disagree, the proposed framework yields an accuracy of $60.4\%$, roughly doubling that obtained by the baseline ($30.5\%$). Analogously, in comparison with the many-to-one baseline, the label agreement is as high as $92.0\%$  whereas an accuracy gap of $15.2\%$ is seen on the dissenting subset with $52.4\%$ of the proposed framework compared to $37.2\%$ of the baseline. 

\setlength\tabcolsep{2.5pt} 
\begin{table}[!b]
	\caption{The overall accuracy of different frameworks on MASS's transition and non-transition subsets. The results are obtained by 1-max CNN base with $Q=1000$ and $P=3$.}
	\footnotesize
	\vspace{-0.2cm}
	\begin{center}
		\begin{tabular}{|>{\arraybackslash}m{0.7in}|>{\centering\arraybackslash}m{0.65in}|>{\centering\arraybackslash}m{0.65in}|>{\centering\arraybackslash}m{0in} @{}m{0pt}@{}}
			\cline{2-3}
			\multicolumn{1}{c|}{} & Non-transition (Size $83.4\%$) & Transition (Size $16.6\%$) &  \parbox{0pt}{\rule{0pt}{2ex+\baselineskip}} \\ [0ex]  	
			\cline{1-3}
			One-to-many  & $89.5$ & $53.3$ &  \parbox{0pt}{\rule{0pt}{0.5ex+\baselineskip}} \\ [0ex]  	
			One-to-one  & $86.8$ & $62.0$ &  \parbox{0pt}{\rule{0pt}{0.5ex+\baselineskip}} \\ [0ex]  	
			Many-to-one  &$88.2$ & $51.1$ &  \parbox{0pt}{\rule{0pt}{0.5ex+\baselineskip}} \\ [0ex]  	
			\cline{1-3}
		\end{tabular}
	\end{center}
	\label{tab:performance_transition_nontransition}
\end{table}

\subsection{Influence of the Number of Convolutional Filters}

In general, more features can be learned by the proposed 1-max CNN with the increasing number of convolutional filters $Q$ and one can expect improvement on the performance. However, influence of $Q$ on the framework's performance is very modest as can be seen from Figure \ref{fig:context_smoothing}. 
For instance, on Sleep-EDF, fixing $P=2$ and multiplicative voting, using $Q=1000$ only brings up $0.5\%$ absolute accuracy gain over the case of $Q=100$ even though the number of filters is ten times larger. A similar finding can also be drawn for MASS ($P=3$) with a modest improvement of $0.4\%$. The slight influence of the number of filters $Q$ suggests that we can maintain a very good performance even with a much smaller network.

\subsection{Benefits of Multimodal Input}

Single-channel EEG has been found prevalent in literature \cite{Koley2012, Kuo2011, Supratak2017, Tsinalis2016, Phan2018c, Phan2018d} mainly due to its simplicity. However, apart from brain activities, sleep also involves eye movements and muscular activities at different levels. For instance Rapid Eye Movement (REM) stage usually associates with rapid eye movements and high muscular activities are usually seen during the Awake stage. As a result, EOG and EMG are valuable additional sources, complementing EEG in multimodal automatic sleep staging systems \cite{Chambon2018, Lajnef2015, Huang2014, Mikkelsen2018, Andreotti2018, Stephansen2017}, not to mention their importance in manual scoring rules \cite{Hobson1969,Iber2007}.

Figure \ref{fig:context_smoothing} reveals and demonstrates the benefit of using EOG and EMG to complement EEG in the proposed framework. Consistent improvements on overall accuracy can be seen on both Sleep-EDF and MASS. Taking MASS for example, averaging over spectrum of $Q$, as compared to the single-channel EEG, coupling EEG and EOG leads to an absolute gain of $4.1\%$ and is further boosted by another $1.1\%$ with the compound of EEG, EOG, and EMG. 

\subsection{The Trade-off Problem with the Output Context Size}

It is straightforward to extend the output context in the proposed framework. Doing so, we are able to increase the number of decisions in an ensemble, which is expected to enhance the classification performance \cite{Hinton2015}. However, extending the output context confronts us with a trade-off problem. A large context weakens the link between the input epoch and the far-away neighbors in the output context. Oftentimes, this deteriorates the prediction decisions on these epochs and, as a consequence, reduces the quality of individual decisions in the ensemble. The low quality of these prediction decisions may outweigh the benefits of the increased cardinality, worsening the performance instead collectively.

\begin{figure} [!t]
		\centering
		\includegraphics[width=1\linewidth]{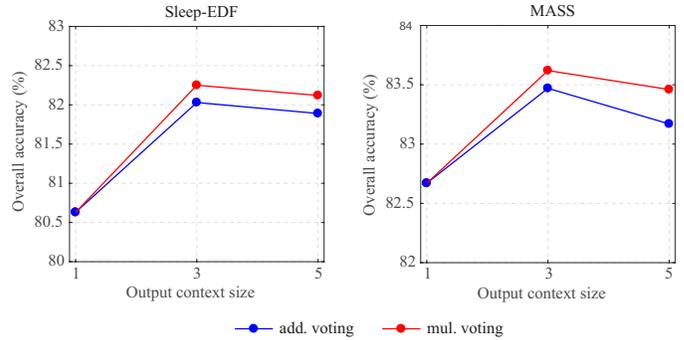}
		\caption{Influence of the output context size to the overall accuracy of the proposed framework. The results obtained with a common $Q=1000$, in addition, $P=2$ (Sleep-EDF) and $P=3$ (MASS).}
		\label{fig:influence_contextsize}
\end{figure}

To support our argument, we increased the output context size to 5 (i.e. $\tau=2$) and repeated the experiment in which we set $Q=1000$ for the 1-max CNN and used $P=2$ for Sleep-EDF and $P=3$ for MASS.
Figure \ref{fig:influence_contextsize} shows the obtained performance alongside those obtained with the output context size of \{1, 3\} (i.e. $\tau = \{0,1\}$). Note that, with the context size of 1, the framework is reduced to the one-to-one baseline framework described in Section \ref{ssec:baseline}. With the context size of $5$ the proposed framework still maintains its superiority over the standard classification setup, however, a graceful degradation compared to the context size of $3$ can be observed. Specifically, the accuracy rates obtained by both additive and multiplicative voting schemes slightly decline by $0.1\%$ on Sleep-EDF while the respective accuracy losses of $0.3\%$ and $0.2\%$ can be seen on MASS.

To remedy the weak links between the input epoch and far-away epochs, one possibility is to combine multiple epochs into the input to form the contextual input. In addition, it would be worth exploring incorporation of long-term context (i.e. in order of dozens of epochs), for example using RNNs as in \cite{Supratak2017,Stephansen2017}. However, a detailed study of the proposed frame work in these many-to-many settings is out of the scope of this article and is left for future work.

\subsection{Multitask vs Ensemble}

\begin{figure} [!t]
		\centering
		\includegraphics[width=1\linewidth]{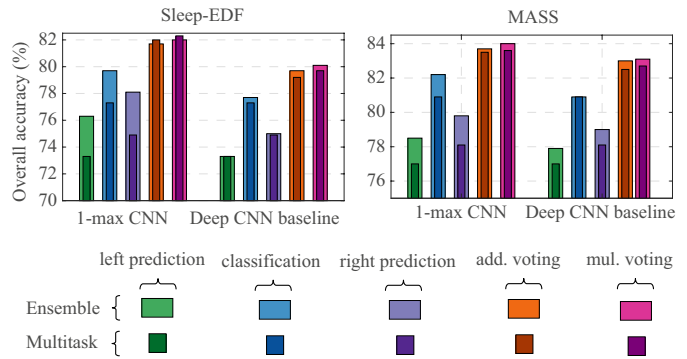}
		\caption{Performance comparison of the proposed multitask 1-max CNN with its equivalent ensemble model. The results are obtained with $Q=1000$, $P=2$ with Sleep-EDF and $P=3$ for MASS.}
		\label{fig:multitask_vs_ensemble}
\end{figure}

To examine the comparability between the proposed multitask framework with its ensemble equivalence, we repeated the experiments with the ensemble consisting of three separate CNNs for individual subtasks: left prediction, classification, and right prediction. We studied both the 1-max CNN and the deep CNN baseline here. Again, we set $Q=1000$, and $P=2$ for Sleep-EDF and $P=3$ for MASS when 1-max CNN was used. The results obtained with the ensemble models and the proposed multitask models are contrasted \emph{vis-\`{a}-vis} in Figure~\ref{fig:multitask_vs_ensemble}.

Our analyses show that the separate CNNs of an ensemble model perform better than its corresponding multitask model on the individual subtasks. This is due to the fact that the multitask model needs to deal with a harder modelling task which combines all the subtasks as a whole. However, after aggregation, their differences become negligible as can be seen over all CNN architectures and datasets. More importantly, on both datasets, the proposed multitask 1-max CNN outperforms the deep CNN baseline in its both forms, namely multitask and ensemble.


\section{Conclusions}
\label{sec:conclusion}

This work introduced a joint classification and prediction formulation wherein a multi-task CNN framework is proposed for automatic sleep staging. Motivated by the dependency nature of sleep epochs, the framework's purpose is to jointly perform classification of an input epoch and prediction of the labels of its neighbors in the context output. While being orthogonal to the widely adopted many-to-one classification scheme relying on contextual input, we argued that the proposed framework avoids the shortcomings experienced by the many-to-one approach, such as the inherent modelling ambiguity and the induced computational overhead due to large contextual input. More importantly, due to multitasking, the framework is able to conveniently produce multiple decisions on a certain epoch thereby forming the reliable final decision via aggregation. We demonstrated the generalizability of the framework on two public datasets, Sleep-EDF and MASS.

\section*{Acknowledgment}
The research was supported by the NIHR Oxford Biomedical Research Centre and Wellcome Trust under Grant 098461/Z/12/Z.

\ifCLASSOPTIONcaptionsoff
  \newpage
\fi



%
\vspace{-0.2cm}
\bibliographystyle{IEEEtran}
\bibliography{bibliography}

\end{document}